\documentclass{article}
\usepackage{amsmath}
\usepackage{microtype}
\usepackage{graphicx}
\usepackage{subfigure}
\usepackage{booktabs} 
\usepackage{diagbox}
\usepackage{hyperref}

\usepackage[accepted]{icml2021}

\begin{document}

\twocolumn[
\icmltitle{ClueGAIN: Application of Transfer Learning On Generative Adversarial Imputation Nets (GAIN)}




\begin{icmlauthorlist}
\icmlauthor{Simiao Zhao}{}

\end{icmlauthorlist}


\vskip 0.3in
]




\begin{abstract}
Many studies have attempted to solve the problem of missing data using various approaches. Among them, Generative Adversarial Imputation Nets (GAIN) was first used to impute data with Generative Adversarial Nets (GAN) and good results were obtained. Subsequent studies have attempted to combine various approaches to address some of its limitations. ClueGAIN is first proposed in this study, which introduces transfer learning into GAIN to solve the problem of poor imputation performance in high missing rate data sets. ClueGAIN can also be used to measure the similarity between data sets to explore their potential connections.
\end{abstract}

\section{Introduction}
\label{intro}
Processing missing data is one of the unavoidable problems in data analysis. Important information can be lost if observation of the missing part is simply discarded, resulting in a systematic difference between incomplete and complete observed data. Therefore, data scientists have done a large amount of work, such as MICE \cite{van2011mice, buuren2000multivariate}, MissForest \cite{stekhoven2012missforest} and DAE \cite{vincent2008extracting},  to find reliable methods to impute missing regions with rational values.

Yoon et al. first proposed Generative Adversarial Imputation Net (GAIN) to impute data Missing Completed At Random (MCAR) \cite{yoon2018gain}. GAIN performs better than the traditional imputation method and does not rely on complete training data.  However, it still has some limitations, mainly from the model structure and the assumptions about data.

Firstly, the simple structure of GAIN is not able to effectively complete the imputation task with high data missing rate. To solve this problem, MisGAN \cite{li2019misgan}, with two pairs of generators and discriminators, Generative Adversarial Multiple Imputation Network (GAMIN) \cite{yoon2020gamin}, which used the confidence prediction method, and the Generative Adversarial Guider Imputation Network (GAGIN) \cite{wang2022gagin} consisting of three models, was proposed.

Secondly, GAIN did not show satisfactory performance when being applied to time series imputation tasks. Therefore, Two-stage GAN \cite{andreini2021two} and  Multivariate Time Series GAN (MTS-GAN) \cite{luo2018multivariate} based on time series imputation were first proposed. End-to-end GAN(E2GAN) \cite{luo2019e2gan} and Inverse Mapping GAN (IMGAN) \cite{wu2022imputing} were subsequently proposed to improve efficiency and performance.

Thirdly, the theory guarantees of GAIN based on MCAR assumption, which is not always true since the lost data may depend on observed variables (MAR assumption) or even unobserved variables (MNAR assumption). Fang et al. extended the theoretical results of GAIN to MAR, and eliminated the need for hint mechanism \cite{fang2022fragmgan}.

Finally, by adjusting loss function and the structure of GAIN, the imputation performance of GAIN can be further improved, and the common problems of GAN itself, such as gradient vanishing of generator and model collapse, can be solved. WGAIN \cite{friedjungova2020missing} introduced wasserstein distance to loss function to solve the problem of model collapse. GRAPE \cite{you2020handling} proposes a graph-based framework for data imputation. PC-GAIN \cite{wang2021pc} and HexaGAN \cite{hwang2019hexagan} introduced unsupervised learning to improve GAIN performance and stability. Conv-GAIN added the structure of the convolutional neural network (CNN) to GAIN \cite{adeli2021convolutional}. 

 In this paper, we proposes ClueGAIN, which combines transfer learning (TL) with GAIN to improve the performance on high missing rate data of GAIN. Although in many cases it may not be possible to obtain complete data that is exactly the same as the missing data to be repaired for training purposes \cite{yoon2018gain}, data that are potentially similar to the target data are not necessarily difficult to obtain. These similar data can provide the model with some prior knowledge, the 'clues', about the target data through TL.
In 1976, Stevo Bozinovski et al. first gave a mathematical and geometrical model of TL \cite{bozinovski1976influence}. Some subsequent studies combined GAN and TL for image classification \cite{cho2017neural,li2018cc}, while others applied TL alone for data imputation \cite{ma2020transfer}. However, TL was never combined with GAN to impute data, and we first propose this idea. In addition, we also discusses the possibility of using ClueGAIN to measure the degree of similarity between multiple data sets. This measure can be applied to look for similarity between biomedical data, such as the similarity between different genes, drugs and proteins.

\section{Problem Formulation}
\label{pf}
Consider a d-dimensional space ${\displaystyle \mathcal{X} = \mathcal{X}_1 \times... \times \mathcal{X}_d}$ . Suppose that $\textbf{X} = (X_1, ..., X_d)$ is a random variable (either
continuous or binary) taking values in $\mathcal{X}$ , whose distribution we will denote $P(\textbf{X})$. Suppose that $\textbf{M} = (M_1, ..., M_d)$ is a random variable taking values in ${\{0, 1\}}^d$. We will call $\textbf{X}$ the data vector, and $\textbf{M}$ the mask vector.
For each $i \in \{1, ..., d\}$ we define a new space $\tilde{\mathcal{{X}}_i} = \mathcal{X}_i \cup {\{*\}}$ where $*$ is simply a point not in any $ \mathcal{X}_i$, representing an unobserved value. Let  $ \tilde{\mathcal{X}} = \tilde{\mathcal{X}}_1 \times ... \times \tilde{\mathcal{{X}}}_d$. We define a new random variable ${\tilde{\textbf{X}}} = (\tilde{X}_
1, ...,\tilde{{X}_d}) \in \tilde{{\mathcal{X}}}$ in the following way:
\begin{equation}
 \tilde{{X}}_i = \begin{cases}
X_i &
\text{if $M_i = 1$}\\
*& \text{otherwise}
\end{cases}
\end{equation}
The distribution is denoted by $P(\tilde{\textbf{X}})$.
$n$ i.i.d. copies of $\tilde{\textbf{X}}$ are realized, denoted $\tilde{x}^1, ..., \tilde{x}^n$ and we define the data set $\mathcal{D} ={\{(\tilde{x}^i, m_i)\}_{i=1}^n}$ \cite{yoon2018gain}.

The transfer learning problem is given in terms of domains and tasks. Given a specific domain, $ {\displaystyle {\mathcal {O}}=\{ {\tilde{\mathcal{{X}}}}, P(\tilde{\textbf{X}}})\} $, a task, denoted by ${\displaystyle {\mathcal {T}}=\{{\mathcal {X}}, h(P(\tilde{\textbf{X}}))\}}$, consists of three components: a d-dimensional space $\mathcal{X}$ mentioned above, a model
${\displaystyle h:{\tilde{\mathcal {X}}}\rightarrow {\mathcal {X}}}$ and a sampler $G$. The model $h$ is used to produce a distribution $h(P(\tilde{\textbf{X}}))$ that is closest to (in the best case the same as) the distribution $P(\textbf{X})$ based on $P(\tilde{\textbf{X}})$, and $P(\tilde{\textbf{X}})$ is decided by the observed data set $\mathcal{D}$. We can therefore sample data from $h(P(\tilde{\textbf{X}}))$ in order to impute the missing values.

Given a source domain ${\displaystyle {\mathcal {O}}_{S}}$, a target domain ${\displaystyle {\mathcal {O}}_{T}}$ and learning task ${\displaystyle {\mathcal {T}}_{T}}$, where ${\displaystyle {\mathcal {O}}_{S}\neq {\mathcal {O}}_{T}}$, transfer learning is used to help improve the learning of the target model ${\displaystyle h_{T}(\cdot)}$ in ${\displaystyle {\mathcal {O}}_{T}}$ using the knowledge in ${\displaystyle {\mathcal {O}}_{S}}$. Furthermore, it is used on measure how close  ${\mathcal {O}}_{S}$ to ${\mathcal {O}}_{T}$ by measuring the contribution of the knowledge in ${\displaystyle {\mathcal {O}}_{S}}$ on learning the target model ${\displaystyle h_{T}(\cdot )}$. Thus, the similarity between the true distribution $P_S(\textbf{X})$ and $P_T(\textbf{X})$ can be further inferred from $P_S(\tilde{\textbf{X}})\in {\mathcal {O}}_{S}$ and $P_T(\tilde{\textbf{X}})\in {\mathcal {O}}_{T} $.

\section{Clue Generative Adversarial Imputation Nets}
In this section, we will go through the overall process of using ClueGAIN to impute missing data and measure the similarity between different data sets. 

\subsection{Data Imputation}
ClueGAIN's data imputation process is divided into two steps, pre-training and fine-tuning. Therefore, the model requires two data sets, the source data set $S$ and the target data set $T$. We assume that the source data set is complete and similar to the target data set $T$.

\subsubsection{pre-training}
ClueGAIN, like GAIN, consists of a generator $G$ and a discriminator $D$. The generator $G$, takes $\tilde{\textbf{X}}$, $\textbf{M}$ and a noise
variable $\textbf{Z}$, as input and output $\Bar{textbf{X}}$, a vector of imputations.
Let $G : \tilde{\mathcal{X}}  \times \{0, 1\}^
d \times [0, 1]^d \rightarrow{\mathcal{X}} $ be a function, and
$\textbf{Z} = (Z_1, ..., Z_d)$ be d-dimensional noise.
We define random variables ${\Bar{\textbf{X}}, \hat{\textbf{X}} \in \mathcal{X}}$ by

\begin{equation}
 \Bar{{\textbf{X}}} = G(\tilde{\textbf{X}},\textbf{M},(\textbf{1} - \textbf{M}) \odot \textbf{Z} ) 
\end{equation}
\begin{equation}
 \hat{{\textbf{X}}} = \textbf{M} \odot \tilde{\textbf{X}}+ (\textbf{1} - \textbf{M}) \odot \Bar{\textbf{X}} 
\end{equation}

The discriminator is a function $D : \mathcal{X} \rightarrow{[0, 1]^d}$ with the $i$-th component of $D(\hat{x})$ corresponding to the probability that the $i$-th component of $\hat{x}$  was observed.

The training algorithm and loss function that discriminator $D$ needs to optimize is the same as that of GAIN \cite{yoon2018gain}. For 
$\mathcal{L}: \{0,1\}^d \times [0,1]^d \times \{0,1\}^d \rightarrow{R}$:
\begin{equation}
\mathcal{L}_D (\textbf{m}, \hat{\textbf{m}}, \textbf{b}) = \sum_{i:b_i = 0} [m_ilog{\hat{m_i}} + (1 - m_i)log{(1-\hat{m_i})}] 
\end{equation}

The training algorithm for $G$ is also the same as that of GAIN \cite{yoon2018gain}. However, the loss function of $G$ is different from GAIN. In the pre-training stage, no matter $(m_i = 1)$ or $(m_i = 0)$, the loss function that G needs to optimize is always $\mathcal{L}: R^d \times R^d \rightarrow{R}$ :
\begin{equation}
\mathcal{L}_G (x_i,{x_i}^*) = \sum_{i=1}^d  L_G(x_i,{x_i}^*)
\end{equation}
where
\begin{equation}
 L_G(x_i,{x_i}^*) = \begin{cases}
(x_i - x_j)^2 & \text{if $x_i$ is continuous}\\
-x_ilog(x_i^*) & \text{if $x_i$ is binary}
\end{cases}
\end{equation}

This is because the source data set $S$ is complete and the masked data ($x_i$ when $m_i = 0$) can be supplied to the generator for training. Therefore, the loss function for $G$ can be just the reconstruction error of $S$. This will not affect the training of discriminator $D$ because the generator still try to minimize difference between the true values and generated values, while the discriminator still discriminates against the value of $m_i$ based on hints and the data generated by generator.

\begin{figure}[ht]
\vskip 0.2in
\begin{center}
\centerline{\includegraphics[width=\columnwidth]{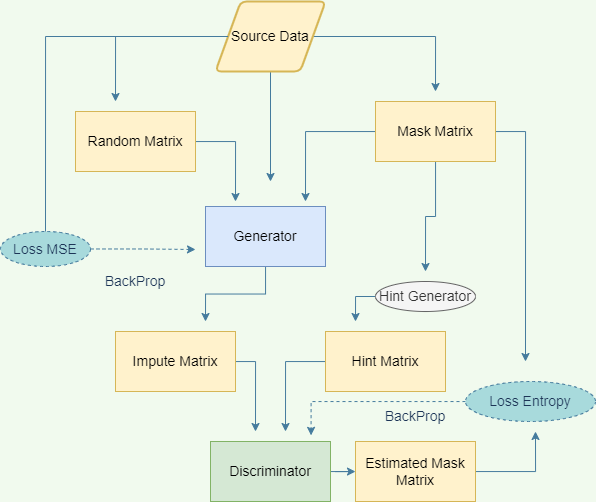}}
\caption{Pre-training Process of ClueGAIN}
\label{p1}
\end{center}
\vskip -0.2in
\end{figure}

The whole process of pre-training is shown in figure \ref{p1}.

\subsubsection{Fine-tuning}
\label{sub}
There are four methods for model fine-tuning. The first method is to retain hidden layers of both generator and discriminator, and only retrain the input and output layers on the target data set (the input and output layers need to be redefined and trained because the feature dimensions of the target and source data set may be different). The algorithm and loss function used for retraining are exactly the same as those used for GAIN \cite{yoon2018gain}. In the second way, part of the hidden layers in the model are preserved (frozen during fine-tuning), and part of the hidden layers, input, and output layers are retrained. The third is to retrain all model parameters, but the pre-trained hidden layer parameters are taken as the initial parameters. The fourth is to redefine the number of network layers and neurons of the new model, and add the pre-trained hidden layers to the hidden layers of the new model. Which approach works best depends on the differences between the target and source data sets. The performance of these four methods are compared and reported in section \ref{er} on the chosen data set. 

After fine-tuning, we can use the generator $G$ to impute data for the target data set $T$.

\subsection{Data Similarity Measurement}
As stated in section \ref{pf}, by measuring how efficient the knowledge in ${\displaystyle {\mathcal {O}}_{S}}$ contributes to the target model ${\displaystyle h_{T}(\cdot )}$, we can get information about the similarity between ${\mathcal {O}}_{S}$ to ${\mathcal {O}}_{T}$, and hence infer the similarity between the true distribution $P_S(\textbf{X})$ and $P_T(\textbf{X})$.

The contribution of source domain to target model can be measured by the difference of performance of ClueGAIN and GAIN. GAIN can be regarded as ClueGAIN without information about the source domain (since there is no pre-training stage). Therefore, if the performance of ClueGAIN is significantly higher than that of GAIN, if other conditions such as the number of layers and neurons in both models are the same, it indicates that the pre-training information of ClueGAIN plays a role in training the target model ${\displaystyle h_{T}(\cdot )}$. 

This approach enables us to compare similarity between multiple data sets. For example, given a target data set and multiple other data sets, using the algorithm \ref{alg:example2}, we can determine which data set is most similar to the target data set.

In biomedical research, it has the potential to help us explore the relationships between different genes, proteins, and drugs. For example, given the attributes of a set of oncogenes, we can determine which genes are similar to them using the algorithm \ref{alg:example2}. See section \ref{dis} for more discussion of applications.

\begin{algorithm}[tb]

   \caption{Data Similarity Measurement Algorithm}
   \label{alg:example2}
\begin{algorithmic}

   \STATE {\bfseries Input:} Mutiple data set $(D_1, ..., D_n)$ and target data set $T$
   \STATE {\bfseries Step 1:} pre-train ClueGAIN on target data set $T$ and save pre-trained parameters.
   \STATE {\bfseries Step 2:} Mask a certain proportion of each data set in ($D_1$, ..., $D_n$).
   \STATE {\bfseries Step 3:} Fine-tune ClueGAIN on each data set in ($D_1$, ..., $D_n$) separately by transferring the pre-trained parameters, impute the masked data, append each performance score to $R_1list$.
   \STATE {\bfseries Step 4:} 
   Train $n$ GAINs on each data set and impute the masked data, append each performance score to a list $R_2list$.
   \STATE {\bfseries Step 5:}\FOR{each $R_1 \in R_1list$ and each $R_2  \in R_2list$}
   \STATE Append $Score_i= {R_1 - R_2}$   to $Scorelist$

   \ENDFOR
    \STATE{\bfseries Step 6:} Find the highest score $Score_i$ in $Scorelist$ 
     \STATE{\bfseries Output:} $D_i$ is most similar to $T$ for $i \in Score_i$

\end{algorithmic}
\vskip 0.1in

\end{algorithm}

\section{Experiments and Results}
\label{er}
\begin{figure}[ht]
\vskip 0.2in
\begin{center}
\centerline{\includegraphics[width=\columnwidth]{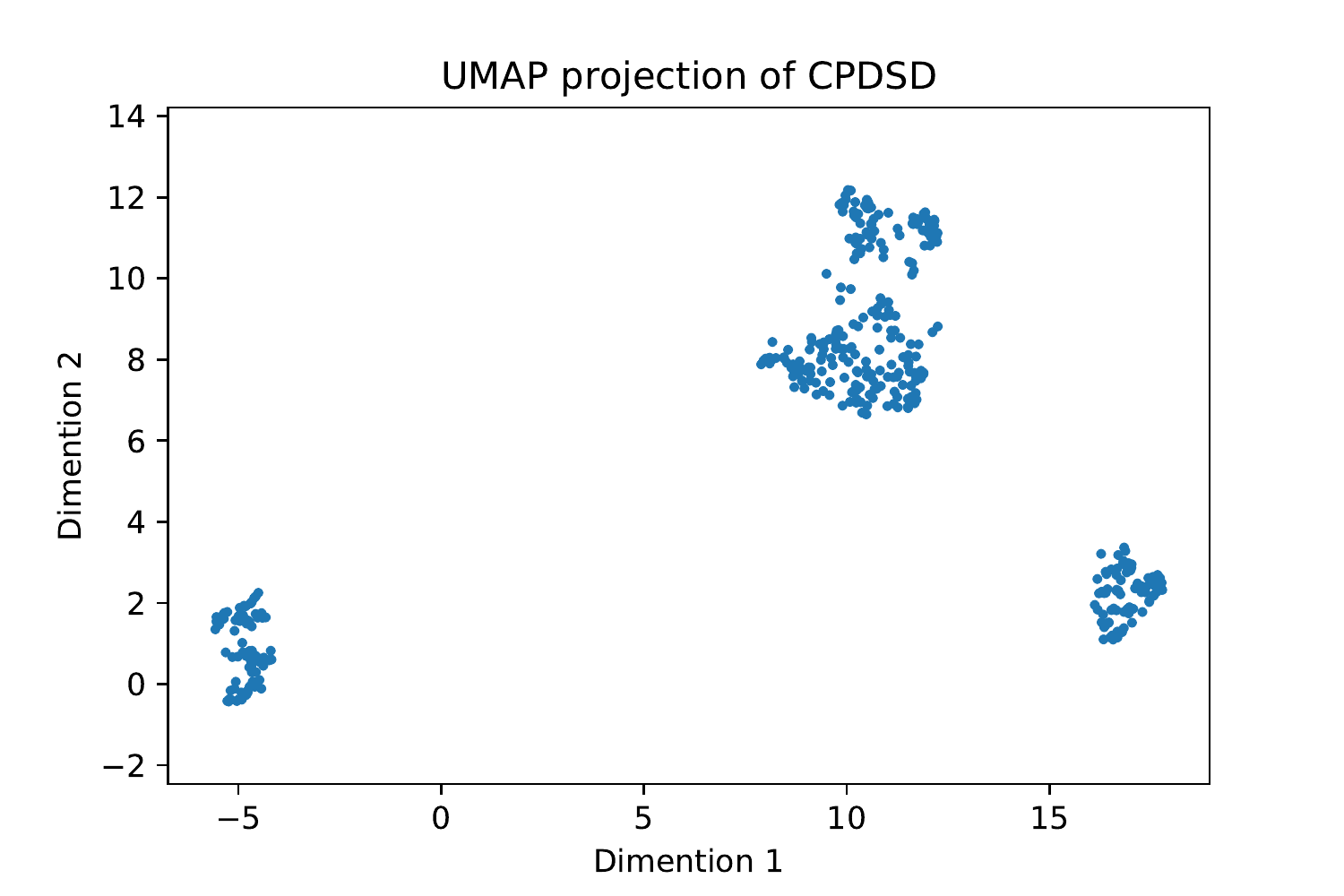}}
\caption{Projection of Cancer Patients DNA Sequence Dataset in Two-dimensional Space}
\label{DNA}
\end{center}
\vskip -0.2in
\end{figure}

\begin{figure}[ht]
\vskip 0.2in
\begin{center}
\centerline{\includegraphics[width=\columnwidth]{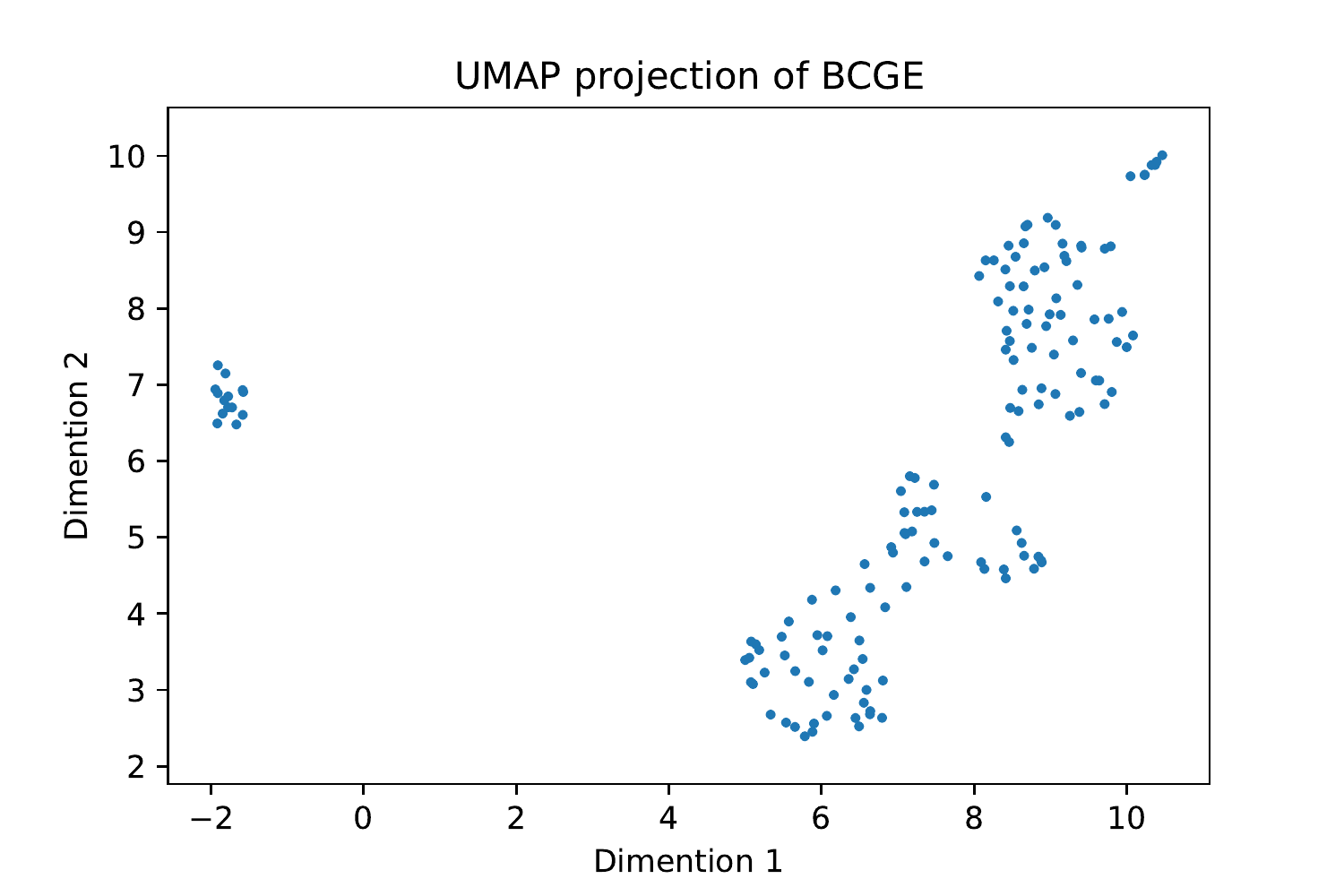}}
\caption{Projection of Breast Cancer Gene Expression in Two-dimensional Space}
\label{BCGE}
\end{center}
\vskip -0.2in
\end{figure}

This section discusses four different experiments and their results. These experiments were carried out mainly on two data sets, Cancer Patients DNA Sequence Dataset (CPDSD) \cite{pd2001} and Breast Cancer Gene Expression - CuMiDa (BCGE) \cite{pd22020}. The CPDSD is a small data set containing 44 genes from 391 patients and a column of five different lass labels. The projection of data in a two-dimensional space is shown in figure \ref{DNA}. BCGE is a large data set containing 54,676 genes from 151 samples and a column of six different class labels. The projection of data on a two-dimensional space is shown in figure \ref{BCGE}.

These two data sets were chosen for two reasons. The first is that the two are potentially related because they are both genetic data sets about cancer patients. Secondly, CPDSD has a small amount of data while BCGE has a large amount. This corresponds to the real situation in the real world, where we may have access to sufficient data of a certain type, for example, a common disease, but data may be lacking or few for a new type, such as a rare or nascent disease.

In the first experiment, we compare the influence of different fine-tuning methods on ClueGAIN's performance with a fixed missing rate (including the performance of GAIN as a benchmark), as mentioned in section \ref{sub}. In the second experiment, we compare the performance of GAIN and ClueGAIN with different missing rates of data sets. The third experiment compares the prediction performance of GAIN and ClueGAIN. A final experiment reveals the feasibility of using the algorithm \ref{alg:example2} to compare the similarity between different data sets. 

There are several points to note:
\begin{itemize}
    \item For the first three experiments, we conduct each experiment ten times as well and report either Root Mean Square Error(RMSE) or Area Under the Receiver Operating Characteristic Curve (AUROC) as the performance metric along with their standard deviations across the 10 experiments.
    \item  For all experiments, pre-training, if any, are carried out on BCGE and CPDSD is used for data imputation.
    \item CPDSD itself has many missing values, which are concentrated in some of the (gene) columns. For the first and second experiments, these missing values will affect the evaluation of the model, so those gene columns with a large number of missing values are discarded in the first two experiments.
\end{itemize}

\subsection{Fine-tuning Methods Comparison}
\label{ftc}
As described in section \ref{sub}, clue-gain has four methods in the fine-tuning stage.
\begin{enumerate}

    \item Directly use all pre-trained hidden layers (and retrained input and output layers) parameters to perform data imputation.\label{1}
    \item Retrain all the hidden layer parameters using pre-trained parameters as initialization.\label{2}
    \item Preserve the pre-trained hidden layer parameters and add them to the newly trained hidden layers and neurons on the target data set for data imputation.\label{3}
    \item Freeze some of the pre-trained hidden layers and retrain the rest.\label{4}   
\end{enumerate}

To complete the comparative experiment, we construct six different neural networks, five ClueGAINs and GAIN as the benchmark. The first, second, and third ClueGAIN correspond to methods \ref{1}, \ref{2}, and \ref{3} above respectively. The remaining two ClueGAINs correspond to the method \ref{4}, where ClueGAIN4 freezes the half of the hidden layers near the input (shallow layers) and ClueGAIN5 freezes the half of the hidden layers near the output (deep layers). This is because we want to explore the effect of freezing different part of the hidden layers on the model. For different types of data, freezing different layers may have different effects on model performance \cite{yosinski2014transferable}. 

The generators and discriminators of all ClueGAINs (and GAIN) except the third ClueGAIN have four hidden layers, each of which has 10 neurons. Multiple hidden layers are added because we need the model to learn general characteristics of the source data set as well as specific characteristics of the target data set, which requires sufficient complexity of the model. The third ClueGAIN has four hidden layers (both generator and discriminator) in the pre-trained stage, and eight hidden layers in the fine-tuning stage. The first four hidden layers are frozen pre-trained hidden layers, and the last four hidden layers are trainable newly-added hidden layers.

Table \ref{sample-table} shows the performance of the six models with missing rate 60$\%$, 70$\%$, 80$\%$ and 90$\%$ on CPDSD, respectively. It can be seen that nearly all ClueGAINs' RMSE are lower than that of GAIN for data with high missing rate. This may attribute to pre-training stage bringing prior knowledge to the model, which can ensure that, at the fine-tuning stage, the model still has a basic 'assumption' of the underlying distribution of the target data sets with high missing rate.

For CPDSD, among the five ClueGAIN models, the best two models in this experiment are ClueGAIN1 and ClueGAIN5, which perform significantly better than GAIN in the data set with high missing rate. They are selected for further comparison in subsequent experiments. The worst model is ClueGAIN2, which sometimes performs worse than GAIN. 
\begin{table*}[t!]
\begin{center}
\begin{small}
\begin{sc}
\caption{Imputation Performance  (Average ± Std of RMSE)}
\label{sample-table}

\begin{tabular}{lcccr}
\toprule
\diagbox[width=10em]{\textbf{Model}}{   \textbf{  Miss Rate}}&
60$\%$& 70$\%$& 80$\%$ &90$\%$ \\ \hline
\textbf{ClueGAIN1} & .1694 (± .0011) & .1709 (± .0013)  & .1929 (± .0032)  & .2305 (± .0177) \\ \hline
ClueGAIN2 & .1771 ( ± .0025)& .1786 (± .0033)  &.1938 (± .0030)  & .2650 (± .0142)\\ \hline
ClueGAIN3 & .1759 ( ± .0022)& .1772 (± .0028)  &.1891 (± .0065) & .2659 (± .0132)  \\ \hline
ClueGAIN4 & .1725 ( ± .0020)& .1754 (± .0025) &.1894 (± .0024) & .2459 (± .0269)  \\ \hline
\textbf{ClueGAIN5} & .1701 ( ± .0010)& .1755 (± .0013) &.1852 (± .0031) & .2170 (± .0165)  \\ \hline
GAIN &  .1762 ( ± .0030)& .1906 (± .0041)  &.2357 (± .0077) & .2611 (± .0130)  \\ \hline
\bottomrule
\end{tabular}
\end{sc}
\end{small}
\vspace*{-5mm}
\end{center}
\end{table*}
\subsection{ClueGAIN and GAIN in Different Missing Rate}
\begin{figure}[ht]
\vskip 0.2in
\begin{center}
\centerline{\includegraphics[width=\columnwidth]{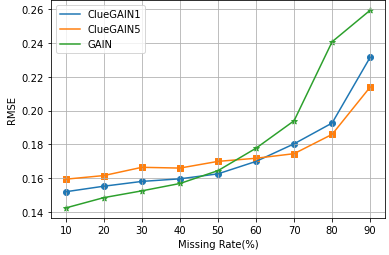}}
\caption{RMSE of ClueGAINs and GAIN vs. Missing Rate on CPDSD}
\label{msr}
\end{center}
\vskip -0.2in
\end{figure}

We now compare the RMSE of two selected ClueGAINs and GAIN at different missing rates of CPDSD. Figure \ref{msr} shows the changes of RMSE of ClueGAIN1, ClueGAIN5 and GAIN with data missing rate. According to the figure, the RMSE of GAIN increases rapidly with increasing missing rate. At about 50$\%$ missing rate, it exceeds RMSE of two ClueGAINs. When comparing two ClueGAINs, ClueGAIN1 has a lower RMSE when the missing rate is less than 60$\%$, but it exceeds ClueGAIN5's RMSE when the missing rate is greater than 60$\%$.

Interestingly, although ClueGAIN5 has the highest RMSE when the missing rate is less than 50$\%$, its imputed data performs best on prediction task. Even when the missing rate is extremely low, it is as good as (or even better than) GAIN whose RMSE is lowest in this range (see section \ref{pp}). This may be because the prior information brought by pre-training helps ClueGAIN5 achieve regularization, reducing the possibility or degree of overfitting. Moreover, RMSE, as a common metric for model performance, has some limitation for its application on data imputation \cite{boursalie2021evaluation}.

\subsection{Prediction Performance}
\label{pp}

\begin{figure}[ht]
\vskip 0.2in
\begin{center}
\centerline{\includegraphics[width=\columnwidth]{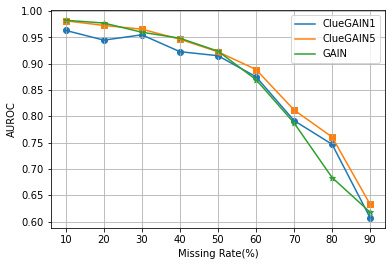}}
\caption{AUROC of ClueGAINs and GAIN vs Missing Rate on CPDSD}
\label{mr}
\end{center}
\vskip -0.55in
\end{figure}

We now compare two selected ClueGAINs against GAIN with
respect to the accuracy of post-imputation prediction. For
this purpose, we use AUROC as a performance measure. To be fair to all methods, we use the same predictive
model (logistic regression) in all cases.
Comparisons are made on CPDSD multi-class classification task and the results are reported in table \ref{t2} and \ref{mr}.

As figure \ref{mr} and table \ref{t2} shows, ClueGAIN5 performs optimally in this experiment. At low missing rate, its AUROC is almost the same as that of GAIN (even slightly higher than that of GAIN), and significantly higher than that of GAIN when missing rate is greater than 50$\%$. It achieves a good result by freezing part of the hidden layers and preserving some prior information about the pre-training data, while leaving some hidden layers to learn the specificity of the target data. In contrast, ClueGAIN1's performance is mediocre, with poor performance at low missing rate and only slightly improved at high missing rate. This result can be attributed to the defect of ClueGAIN1's fine-tuning method, which retains all the hidden layers of pre-training and only retrains the input and output layers. As a result, there are not enough hidden neurons in the model to learn the features that are unique to the target data, which leads to the model being too 'rigid' and unable to fit the target data distribution well.  
\begin{table*}[t!]
\begin{center}
\begin{small}
\begin{sc}
\caption{Prediction Performance at High Miss Rate (Average ± Std of AUROC)}
\label{t2}

\begin{tabular}{lcccr}
\toprule
\diagbox[width=10em]{\textbf{Model}}{   \textbf{  Miss Rate}}&
60$\%$& 70$\%$& 80$\%$ & 90$\%$ \\ \hline
ClueGAIN1 & .8754 (± .0106) & .7920 (± .0189)  & .7469 (± .0143)  & .6070 (± .0227) \\ \hline

ClueGAIN5 & .8891 ( ± .0161)& .8122 (± .0130) &.7607 (± .0141) & .6327 (± .0270)  \\ \hline
GAIN &  .8699 ( ± .0119)& .7875 (± .0073)  &.6836 (± .0149) & .6177 (± .0208)  \\ \hline
\bottomrule
\end{tabular}
\end{sc}
\end{small}
\vspace*{-5mm}
\end{center}
\end{table*}

\subsection{Data Set Similarity Measurement}
The purpose of this experiment is to show the feasibility of using algorithm \ref{alg:example2} for estimating the similarity of data. We introduce various UCI datasets \cite{WinNT2013}, and use the algorithm \ref{alg:example2} to calculate the similarity of BCGE with CPDSD, Spam, Breast, and Letter, respectively. Intuitively, BCGE should be unrelated to the Spam, Breast and Letter data sets.

Table \ref{t3} records how much the RMSE of ClueGAIN5 is lower than that of GAIN on each data set at 80$\%$ missing rate, which used as the score for similarity measurement. As shown in the table, The score of CPDSD is the highest, and the score of the other three data sets is significantly lower and close to zero, which is in line with our expectation that CPDSD is most similar to BCGE.

\begin{table*}[t!]
\begin{center}
\begin{small}
\begin{sc}

\caption{Similarity Scores with BCGE (Average ± Std of $GAIN_{RMSE} - ClueGAIN5_{RMSE}$)}
\label{t3}

\begin{tabular}{lcccr}
\toprule
\hline

\textbf{Data Sets} & CPDSD&  Spam& Breast & Letter \\
\hline

Score & .0691(±.0118) & .0008(±.0002) & .0071(±.0023) & .0037(±.0021)  \\
\hline
\bottomrule
\end{tabular}
\end{sc}
\end{small}
\vspace*{-5mm}
\end{center}
\end{table*}

\section{Discussion}
\subsection{Novelty and Contribution}
\label{dis}
There are two main innovations in this study. The first innovation is that we proposed ClueGAIN, which combined transfer learning with GAIN for the first time, and better performance is achieved on high missing rate data sets. The success of this combination can be extended to other variations of GAIN mentioned in section \ref{intro}. These variants have better performance or more applications than GAIN, and transfer learning may bring further improvement. More importantly, it also provides another idea for data imputation, that is, to search for complete data sets with high similarity to the target data set. A telling example is that in the early days of the COV-19 pandemic, there was a large number of missing values in patient data, either because of government controls or because of the difficulty of data collection. However, the data set for common pneumonia may be relatively sufficient and ClueGAIN pre-trained on it may be able to impute the data for COV-19 patients well.

The second innovation is a new way to measure the similarity of different data sets. This similarity may indicate the underlying true distribution correlation between the two data sets. While traditional methods are usually based on statistics or mathematics, the method in this study is based on computational methods, which can be applied to two sets of data with large size differences. This approach has great potential for biomedical applications. For example, if two seemingly unrelated protein expression data sets can provide useful prior information on the missing data imputation task of each other, which probably means that there are some potential connections between them, such as overlapping genes regulating their expression, etc.
\subsection{Limitations and Future Studies}
There are still many limitations in this study that need to be solved and improved in future research.

The first limitation is that ClueGAIN may not able to outperform GAIN on a low missing rate data set according to experiments. However, as mentioned in section \ref{dis}, future studies can try to combine those well-behaved GAIN variants with transfer learning, for example HexaGAN \cite{hwang2019hexagan}, which claimed better and more stable performance on specific data sets than GAIN.

The second is that ClueGAIN relies too much on finding 'similar' data sets. On the one hand, there are real situations where we can never find a data set that is similar to the target data set. On the other hand, the 'similarity' between the pre-training data set and the target data set is a general intuition in this study, and there is no clear mathematical definition, which may cause that two seemingly similar data sets found is not really similar. Future studies can try to reduce the dependence of ClueGAIN on complete, 'similar' data sets. Also, using mathematical or statistical methods to measure the similarity of the two sets of data sets in advance, so as to prevent the failure of pre-training due to the large difference between the source data set and the target data set, will be helpful.

The third limitation is the high time complexity of the data similarity comparison algorithm. Given $n$ data sets, the algorithm needs to train $2n+1$ models, which is not ideal for large data sets. Future research can try to reduce the complexity of the algorithm.

In addition to improving the above three limitations, future research can focus on two areas. One is to pre-train ClueGAIN on a large number of different data sets. Given ClueGAIN's good performance with just a few data sets for parameter pre-training, this may lead to excellent performance just like BERT in the field of natural language processing \cite{devlin2018bert}. The second is to explore the reasons why different fine-tuning methods affect the performance of the model. In the field of image processing, the research of Jason Yosinski et al. \cite{yosinski2014transferable} has gradually explored the causes and consequences of these effects. However, in the field of data imputation, they are not yet clear and deserve further study.
\section{Conclusion}
In this study, ClueGAIN was proposed based on GAIN. With the idea of layer-based transfer learning, ClueGAIN contains two stages, pre-training and fine-tuning. Through pre-training on a similar data set, prior knowledge of the target data set can be obtained, so as to better complete the imputation task of high missing rate data set. Our experiments proved that ClueGAIN performes better than GAIN in the data imputation task for data set with a high missing rate. In addition, ClueGAIN can be used to measure the similarity between different data sets, which may help to discover potential correlations. However, ClueGAIN still has some limitations, and there is still a lot of room to explore the application and performance of transfer learning combined with GAIN in data imputation.


\bibliography{main}
\bibliographystyle{icml2021}


\end{document}